\RequirePackage{fix-cm}
\documentclass{svjour3}                     % onecolumn (standard format)
\smartqed  % flush right qed marks, e.g. at end of proof
\usepackage{amsmath}
\usepackage{mathtools, cuted}
\usepackage{setspace}
\usepackage{flushend, cuted}
\usepackage{lipsum}
\usepackage{lingmacros}
\usepackage{tree-dvips}
\usepackage{varioref}
\usepackage{colortbl}
\usepackage{url}
\usepackage{hyperref}

\usepackage{graphicx}
\usepackage{epstopdf}
\usepackage{color,soul}
\usepackage{amsfonts}
\usepackage{subfigure}
\usepackage{multicol}
\usepackage{multirow}
\usepackage{hhline}
\usepackage{amssymb}
\usepackage{newlfont}
\usepackage{mathrsfs}
\usepackage{mathtools}
\usepackage{mathrsfs}
\usepackage{cite}
\usepackage{multirow}
\usepackage{nomencl}   
\usepackage{fancyhdr}
\usepackage{caption}
\usepackage{slashbox}
\usepackage{float}
\usepackage{placeins}
\usepackage{tabularx}
\usepackage[table]{xcolor}
\usepackage[linesnumbered,ruled]{algorithm2e}
\SetKwRepeat{Do}{do}{while}%

\DeclareRobustCommand{\frac}[3][0pt]{%
	{\begingroup\hspace{#1}#2\hspace{#1}\endgroup\over\hspace{#1}#3\hspace{#1}}}

\begin{document}

\title{Fast and Robust Localization of Surgical Array using Kalman Filter%\thanks{Grants or other notes
}


\author{Md Ashikuzzaman*         \and
        Noushin Jafarpisheh*  \and
        Sunil Rottoo  \and
        Pierre Brisson \and
        Hassan Rivaz
}


\institute{Md Ashikuzzaman \at
              Department of Electrical and Computer Engineering, Concordia University,   \\
              1455 boul. De Maisonneuve O,\\
              Montreal (Quebec), H3G 1M8, Canada\\
              Tel.: +1 (514) 848-2424 ext.716\\
              \email{m\_ashiku@encs.concordia.ca} \\
              *Equal contribution, order of appearance is determined by flipping a coin. %  \\
}

\date{Received: date / Accepted: date}

\maketitle

\begin{abstract}
Intraoperative tracking of surgical instruments is an inevitable task of computer-assisted surgery. An optical tracking system often fails to precisely reconstruct the dynamic location and pose of a surgical tool due to the acquisition noise and measurement variance. Embedding a Kalman Filter (KF) or any of its extensions such as extended and unscented Kalman filters with the optical tracker resolves this issue by reducing the estimation variance and regularizing the temporal behavior. However, the current rigid-body KF implementations are computationally burdensome and hence, takes long execution time which hinders real-time surgical tracking. This paper introduces a fast and computationally efficient implementation of linear KF to improve the measurement accuracy of an optical tracking system with high temporal resolution. Instead of the surgical tool as a whole, our KF framework tracks each individual fiducial mounted on it using a Newtonian model. In addition to simulated dataset, we validate our technique against real data obtained from a high frame-rate commercial optical tracking system. The proposed KF framework substantially stabilizes the tracking behavior in all of our experiments and reduces the mean-squared error (MSE) from the order of $10^{-2}$~$mm^{2}$ to $10^{-4}$~$mm^{2}$.
\end{abstract}

\section{Introduction}

Kalman Filter (KF) refers to a recursive algorithm which minimizes Mean Squared Error (MSE) and refines the noisy measurements of a system through two stages: prediction and correction~\cite{Welch1995}. Since 1960, when KF was proposed, it has extensively been used in data fusion, tracking and prediction in numerous fields. However, one of the main limitations of KF is that it can be only applied in linear systems. As a consequence, two notable extensions of KF called the Extended Kalman Filter (EKF)~\cite{Jazwinski1970} and Unscented Kalman Filter (UKF)~\cite{Julier1997} have been proposed. EKF linearizes the system under consideration around the operating point and then feeds to the KF. EKF has widely been used in robotics~\cite{Li2020, Aissa2017, Ma2019} and unmanned aerial vehicle~\cite{Prevost2007}. However, EKF is inherently limited by its computational  complexity and long execution time originating from the linearization step. To resolve this important drawback of EKF, researchers have introduced UKF. Instead of linearization, UKF tackles the nonlinearity issue using an unscented transform, where the nonlinear system transforms to a probability distribution function. This innovation empowers UKF to handle a non-linear system faster than EKF and allow real-time computation. Thus far, UKF has been incorporated in a variety of applications namely orientation tracking~\cite{Kraft2003}, mobile robot controlling~\cite{Xu2020}, aerodynamic parameter estimation~\cite{Chowdhary2010} and spacecraft attitude estimation~\cite{VanDyke2004}. 

In recent years, computer aided applications including surgical tracking have emerged rapidly. Optical tracking of the surgical tools often guides the clinicians to perform high-precision surgical procedures~\cite{Elfring2010}. Infra-red emitting diodes, commonly known as active fiducials, are embedded on the surgical tools to be used as reference points to estimate the location and orientation of the tool. However, the noisy observations often leads to an inaccurate estimation of the instrument’s pose which can increase surgical errors. Numerous investigators have employed EKF and UKF to reduce the tracking error by taking the expected noise statistics and temporal constancy into account. Translational velocity and acceleration, and angular velocity are used as the state variables in~\cite{Dorfmuller-Ulhaas2007} to devise an EKF model. Taylor series as well as Rodrigues formula~\cite{Dai2015} have been used to linearize the model and prepare for Kalman filtering. Although encouraging results have been reported, the mathematical formulation is computationally burdening. Hence, Vaccarella \textit{et al.}~\cite{Vaccarella2013} have used UKF to track surgical navigation. A quaternion-based model using translation, linear velocity, quaternion, and angular velocity as the state variables has been adapted to avoid matrix singularity problem that originates from using Euler angles in rotation tracking. Linear acceleration has been added as a state variable in~\cite{Enayati2015} for surgical tracking. 

Although the aforementioned works have reported promising tracking results, EKF and UKF require long execution times and are not suitable for high frame-rate optical tracking applications. Multi-camera optical tracking system to assist total knee arthroplasty (TKA) is such an application which operates at a temporal resolution of approximately $200$ fps. A robust and accurate localization of surgical array is of paramount importance in TKA. However, tracking data obtained by the current system yields high sensitivity to noise. Inspired by previous works, our aim is to combine KF with the existing scheme to increase tracking and localization accuracy of the system. However, the extended and unscented implementations of KF might not be suitable for the system under discussion. Therefore, instead of the whole array as a rigid body, this paper proposes to track each fiducial on the surgical array individually, taking a linear KF into account. This simplification has been possible due to the availability of sufficient temporal information obtained from the high frame-rate system. The advantage of the proposed technique is twofold. First, being fast and computationally light, this framework is compatible with the high frame-rate optical tracking system. Second, any unusual phenomena such as occlusion of a particular fiducial can easily be detected by this scheme since the temporal behavior of each fiducial is assessed individually. We have validated the proposed technique against one simulated and three real datasets obtained using an optical tracking system.

The rest of this paper is organized as follows. In Section 2, we describe the mathematical model pertaining the proposed KF framework. Section 3 describes the experimental setup and data acquisition protocols. In  Section 4, the qualitative and quantitative assessments of the proposed method are provided. Section 5 presents a brief discussion of our findings along with concluding remarks.   
         
\section{Methods}
Let $z_{k}=\begin{bmatrix} p_{x,k} & p_{y,k} & p_{z,k} \end{bmatrix}^{T}$,\; $k \in \{1,2,3,\dots,n\}$ denote the position measurement of the center of a fiducial at time $k$. We assume that $z_{k}$ is corrupted with anisotropic Gaussian noise. Our purpose is to exploit the expected noise statistics of the measured data to minimize the measurement noise and stabilize the temporal tracking using KF.

\subsection{State variables and update equations}
We consider $3D$ components of translation \;$\pmb{t}=\begin{bmatrix} t_{x} & t_{y} & t_{z} \end{bmatrix}^{T}$, velocity \;$\pmb{v}=\begin{bmatrix} v_{x} & v_{y} & v_{z} \end{bmatrix}^{T}$ and acceleration \;$\pmb{a}=\begin{bmatrix} a_{x} & a_{y} & a_{z} \end{bmatrix}^{T}$ as our state variables. Since we assume a constant acceleration motion model, the update equations for the state variables are:

\begin{equation}
\pmb{t}_{k} = \pmb{t}_{k-1} + \pmb{v}_{k-1}\Delta t + \frac{1}{2}\pmb{a}_{k-1}\Delta t^{2}
\label{eq:trans_vec_update}
\end{equation}

\begin{equation}
\pmb{v}_{k} = \pmb{v}_{k-1} + \pmb{a}_{k-1}\Delta t
\label{eq:vel_update}
\end{equation}

\begin{equation}
\pmb{a}_{k} = \pmb{a}_{k-1}
\label{eq:acc_update}
\end{equation}

\noindent
where $\Delta t$ denotes the time interval. 

\subsection{Kalman filter pipeline}
The Kalman filter consists of two steps. The first step predicts the current state and state covariance matrix based on the estimates of the previous time step and the state update model. The second step takes the actual measurement into account to refine the predictions made in the first step.
The state update model described earlier obtains $x_{k}^{-}$, the prior estimate of the current state, using the following linear equation:       
\begin{equation}
x_{k}^{-} = Ax_{k-1}
\label{eq:state update}
\end{equation}

\noindent
where $x_{k-1} = \begin{bmatrix} t_{x} & v_{x} & a_{x} & t_{y} & v_{y} & a_{y} & t_{z} & v_{z} & a_{z} \end{bmatrix}^T$ denotes the posterior state estimate of the previous time step. $A$ describes the motion model and is defined as follows:

\begin{equation}
A=\begin{bmatrix}
1 & \Delta t & \frac{1}{2}\Delta t^{2} & 0 & 0 & 0 & 0 & 0 & 0\\
0 & 1 & \Delta t & 0 & 0 & 0 & 0 & 0 & 0\\
0 & 0 & 1 & 0 & 0 & 0 & 0 & 0 & 0\\
0 & 0 & 0 & 1 & \Delta t & \frac{1}{2}\Delta t^{2} & 0 & 0 & 0\\
0 & 0 & 0 & 0 & 1 & \Delta t & 0 & 0 & 0\\
0 & 0 & 0 & 0 & 0 & 1 & 0 & 0 & 0\\
0 & 0 & 0 & 0 & 0 & 0 & 1 & \Delta t & \frac{1}{2}\Delta t^{2}\\
0 & 0 & 0 & 0 & 0 & 0 & 0 & 1 & \Delta t\\
0 & 0 & 0 & 0 & 0 & 0 & 0 & 0 & 1\\    
\end{bmatrix}
\label{eq:state_mat}
\end{equation}

The prior estimate of the state noise covariance matrix $P_{k}^{-}$ is obtained as follows:
\begin{equation}
P_{k}^{-} = AP_{k-1}A^{T} + Q
\label{eq: cov_update}
\end{equation}
\noindent
where $P_{k-1}$ refers to the posterior estimate of the state covariance matrix obtained from the previous time step. $Q$ denotes the process covariance matrix. Taking $P_{k}^{-}$ into account, the Kalman gain $K_{k}$ is calculated as follows: 

\begin{equation}
K_{k} = P_{k}^{-}H^{T}(HP_{k}^{-}H^{T} + R)^{-1}
\label{eq: kal_gain}
\end{equation}

\noindent
where $R$ stands for the measurement covariance matrix. $H$ obtains a prediction of the measurement using the state prediction and is defined as:

\begin{equation}
H=\begin{bmatrix}
1 & 0 & 0 & 0 & 0 & 0 & 0 & 0 & 0\\
0 & 0 & 0 & 1 & 0 & 0 & 0 & 0 & 0\\
0 & 0 & 0 & 0 & 0 & 0 & 1 & 0 & 0\\    
\end{bmatrix}
\label{eq:H_mat}
\end{equation}
Once the priori estimates and the Kalman gain are calculated, the actual measurement is incorporated to fine-tune the state prediction. We use the difference between predicted and actual measurements to obtain the posterior state estimate according to the following equation:
\begin{equation}
x_{k} = x_{k}^{-} + K_{k}(z_{k}-Hx_{k}^{-})
\label{eq: state_post}
\end{equation}
The posterior estimate of the state covariance matrix is calculated as follows: 
\begin{equation}
P_{k} = (I-K_{k}H)P_{k}^{-}
\label{eq: cov_post}
\end{equation}
The refined measurement $z_{k,r}$ is calculated using:
\begin{equation}
z_{k,r} = Hx_{k}
\label{eq: refine_measurement}
\end{equation}

\noindent
The workflow is outlined in Algorithm 1.

\begin{algorithm}[tb]
	\SetKwInOut{Input}{Input}
	\SetKwInOut{Output}{Output}
	\KwIn{Initial state covariance matrix $P_{0}$, process covariance matrix $Q$ and measurement covariance matrix $R$}
	\KwOut{Refined measurements of fiducial positions}
	
	\For{all fiducials and $k \in \{1,2,3,\dots,n\}$}{
		Estimate $x_{k}^{-}$: Priori state estimate using process model matrix $A$ and previous state estimate $x_{k-1}$\;
		Calculate $P_{k}^{-}$: Priori estimate of the state covariance matrix using $A$, $Q$ and previous state covariance estimate $P_{k-1}$\;
		Compute $K_{k}$: Kalman gain using $P_{k}^{-}$ and $R$\;
		Estimate $x_{k}$: Posterior state estimate using $K_{k}$, $x_{k}^{-}$ and measurement $z_{k}$\;
		Calculate $P_{k}$: Posterior state covariance estimate using $K_{k}$ and $P_{k}^{-}$\;
		Extract the refined position measurements from $x_{k}$
	}
	\caption{Workflow of the proposed Kalman filtering algorithm}
\end{algorithm}

\section{Experimental setup and data acquisition}
In this section, we first describe the simulation experiment conducted to generate synthetic dataset. Then we describe the experimental setup and data collection protocol.

\subsection{Simulated dataset}
We designed a surgical array with four coplanar fiducials. The marker geometry is defined by the mutual distances among the centers of the fiducials. The distances from fiducial 1 to 2, 2 to 3, 3 to 4 and 4 to 1 are $481.04$ mm, $121.66$ mm, $28.28$ mm, and $382.88$ mm, respectively. The motion model of the marker as well as the fiducials is stated below.

Let $X_{0} = \begin{bmatrix} x_{f} & y_{f} & z_{f} \end{bmatrix}^{T}$ denote the initial position of the center of a fiducial. $X_{k}$, the position of the fiducial at time sample $k$, can be calculated by $X_{k} = \pmb{T}_{k} \begin{bmatrix} X_{0}^{T} & 1 \end{bmatrix}^{T}$ where $\pmb{T}_{k} \in \mathbb{R}^{4 \times 4}$ refers to a transformation matrix explaining the pose of the marker at time $k$. $\pmb{T}_{k}$ is defined as follows:  

\begin{equation}
\pmb{T}_{k}=\begin{bmatrix}
\pmb{R}_{k} & \pmb{t}_{k} \\
O & 1  
\end{bmatrix}
\label{eq:trans_mat}
\end{equation}

\noindent
where $\pmb{R}_{k} \in \mathbb{R}^{3 \times 3}$ denotes a $3D$ rotation matrix. $O \in \mathbb{R}^{1 \times 3}$ refers to a zero vector. We consider a constant acceleration translation model. Therefore, the translation update model is governed by Eqs.~\ref{eq:trans_vec_update}-\ref{eq:acc_update}. Considering the time interval $\Delta t$ to be tiny, the rotation matrix $\pmb{R}_{k}$ at time sample $k$ is calculated using the following forward kinematics:

\begin{equation}
\pmb{R}_{k} \approx \pmb{R}_{k-1} + \Delta t \pmb{R^{'}}_{k-1}
\label{eq:rot_mat_Taylor}
\end{equation}

\noindent
where $\pmb{R^{'}}_{k-1}$ refers to the temporal derivative of the rotation matrix at time sample $k-1$ which is defined as follows:

\begin{equation}
\pmb{R^{'}}_{k-1} = \pmb{\omega}\pmb{R}_{k-1}
\label{eq:rot_der_calc}
\end{equation}

\noindent
where $\pmb{\omega} \in \mathbb{R}^{3 \times 3}$ denotes a skew symmetric matrix which is obtained from the angular velocity vector $\omega = \begin{bmatrix} \omega_{x} & \omega_{y} & \omega_{z} \end{bmatrix}$ and defined as:

\begin{equation}
\pmb{w}=\begin{bmatrix}
0 & -\omega_{z} & \omega_{y} \\
\omega_{z} & 0 & -\omega_{x} \\
-\omega_{y} & \omega_{x} & 0 
\end{bmatrix}
\label{eq:omega_skew}
\end{equation}  

We assumed a constant angular velocity vector $\begin{bmatrix} 0.001 & 0.001 & 0.001 \end{bmatrix}$ for our simulation experiment. The initial translational velocity vector was considered to be $\begin{bmatrix} 0 & 0 & 0 \end{bmatrix}$ whereas the translational acceleration vector was set to $\begin{bmatrix} 0.1 & 0.1 & 0.1 \end{bmatrix}$. The initial positions of fiducials 1 to 4 were set to $\begin{bmatrix} -180 & 180 & 1230 \end{bmatrix}^{T}$, $\begin{bmatrix} 170 & -150 & 1230 \end{bmatrix}^{T}$, $\begin{bmatrix} 50 & -130 & 1230 \end{bmatrix}^{T}$ and $\begin{bmatrix} 70 & -110 & 1230 \end{bmatrix}^{T}$, respectively. We choose this order of initial positions to imitate the experimental set-up. Considering 200 temporal samples per second, the fiducial positions for 5 seconds were obtained. Once the ground truth fiducial positions are generated, anisotropic zero-mean random Gaussian noise with a variance of $0.15$~$mm^{2}$ in X and Y directions was added to obtain noisy measurement data. To emulate the real scenario, the noise variance in Z direction was considered to be $40\%$ higher than the other two directions.

\begin{figure*}
	%\DeclareGraphicsExtensions{.eps}
	\centering
	\subfigure[multi-camera system]{{\includegraphics[width=.3\textwidth]{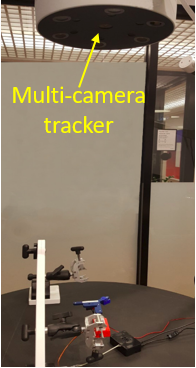}}}%
	\qquad
	\subfigure[Surgical array]{{\includegraphics[width=.41\textwidth]{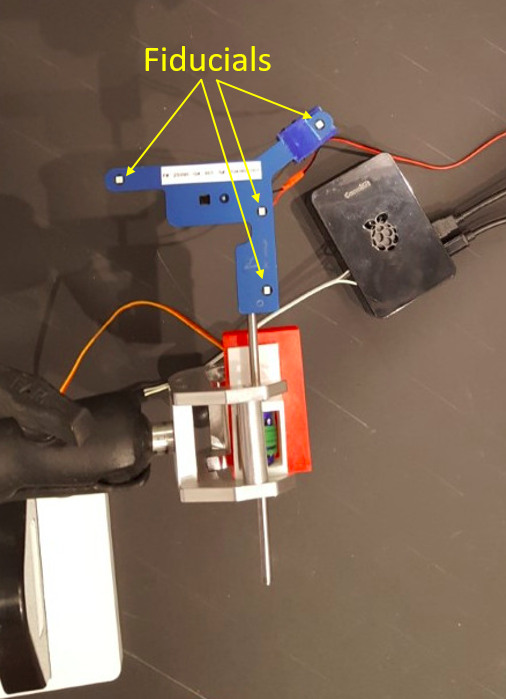}}}
	\caption{Experimental set up of a multi-camera system}
	\label{exp_pics_1}
\end{figure*}

\subsection{Real datasets}
The experimental setup includes a multi-camera optical tracking device which is mounted horizontally on a stable arm above a sturdy table (see Fig.~\ref{exp_pics_1}(a)). The tracker is connected to a host computer that is used to operate the tracker and collect the data. The experiment uses a single medical array with four calibrated fiducials (see Fig.~\ref{exp_pics_1}(b)). The fiducials emit high-intensity near-infrared light with $850$ nm wavelength. The tracker is equipped with infrared filters and reconstructs the 3D position of each fiducial using a standard linear triangulation stereo method. The position of the array is determined by matching the reconstructed 3D points against the calibrated geometry of the array. The resulting pose and 3D points are sent to the host at a rate of approximately $200$ fps. The host stores the data in a relational database for later analysis.

We ran 3 recording sessions under different conditions. For the first recording, the array was left in a static position relative to the tracker and was positioned slightly off-center of the tracker's optical axis. The array was recorded for about $2$ minutes, providing $23,452$ data points.

The second recording was performed on an array undergoing rotational and translation motion, emulating the scenario of a real surgical environment. The recording lasted for about $30$ seconds and obtained $6000$ temporal data points.

The third dataset was acquired from a static array where one of the fiducials was partially occluded using a $220$ GRIT diffuser (Edmund Optics, Barrington, USA). This experiment aimed at imitating the scenario of a real surgical room where fiducials are often blocked by translucent materials such as a drop of blood. The experiment was repeated 4 times, each time occluding one of the 4 fiducials. The array of interest was recorded for about 2 minutes and $22,897$ data points were obtained.

\section{Results}
We examine qualitative and quantitative tracking performance of the proposed Kalman filtering scheme by employing simulation and real datasets obtained from multi-camera system. We use MSE and error variance for the quantitative analysis, with MSE defined as:

\begin{equation}
\textrm{MSE}=\frac{\sum_{k=1}^{n} (p_{r,k}-p_{g,k})^{2}}{n}  
\end{equation}       

\noindent
where $p_{r,k}$ and $p_{g,k}$ denote measured and ground truth positions, respectively. We obtain $Q$ by calculating the covariance matrices corresponding to zero-mean Gaussian random noise with variances of $0.001$~$mm^{2}$ and $0.01$~$mm^{2}$ for simulation and real datasets, respectively. For all datasets, $R$ is obtained by computing the covariance of anisotropic zero-mean Gaussian random noise with a variance of $0.15$~$mm^{2}$ in $x$ and $y$ directions and $0.21$~$mm^{2}$ in the $z$ direction.     

\subsection{Simulated data}
\begin{figure*}
	%\DeclareGraphicsExtensions{.eps}
	\centering
	\subfigure[Fiducial 1, $x$ direction]{{\includegraphics[width=.33\textwidth]{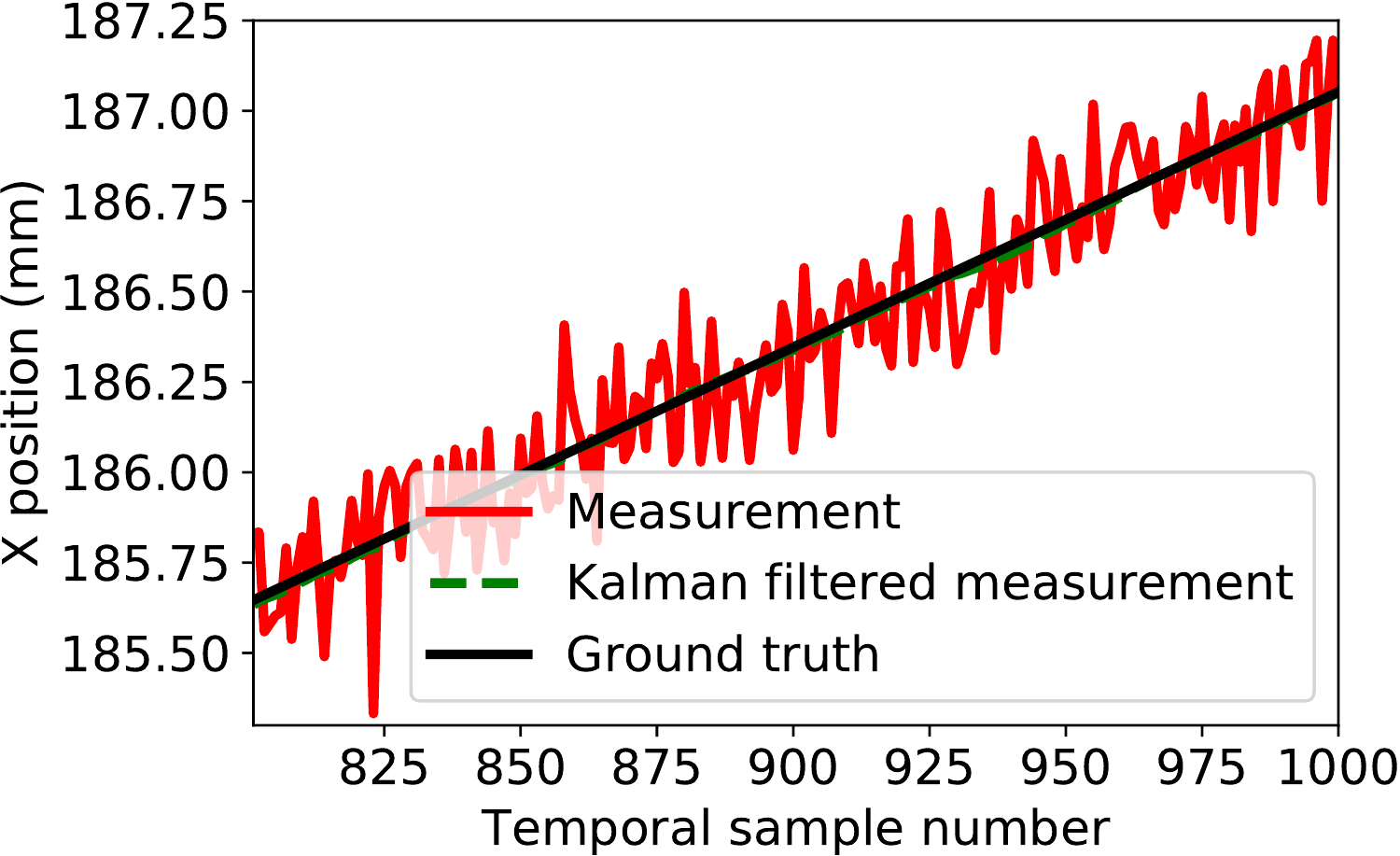}}}%
	\subfigure[Fiducial 1, $y$ direction]{{\includegraphics[width=.33\textwidth]{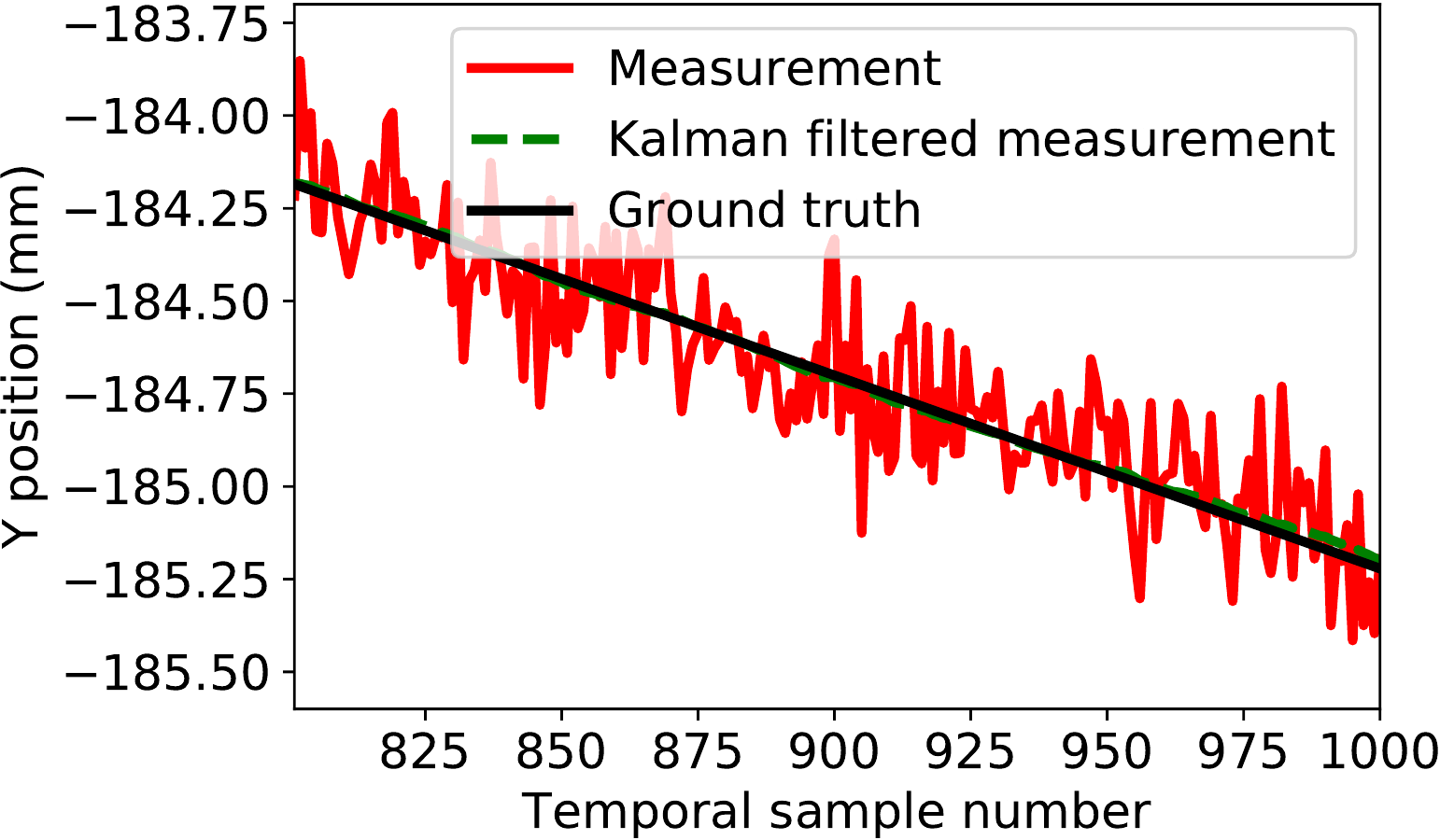}}}%
	\subfigure[Fiducial 1, $z$ direction]{{\includegraphics[width=.33\textwidth]{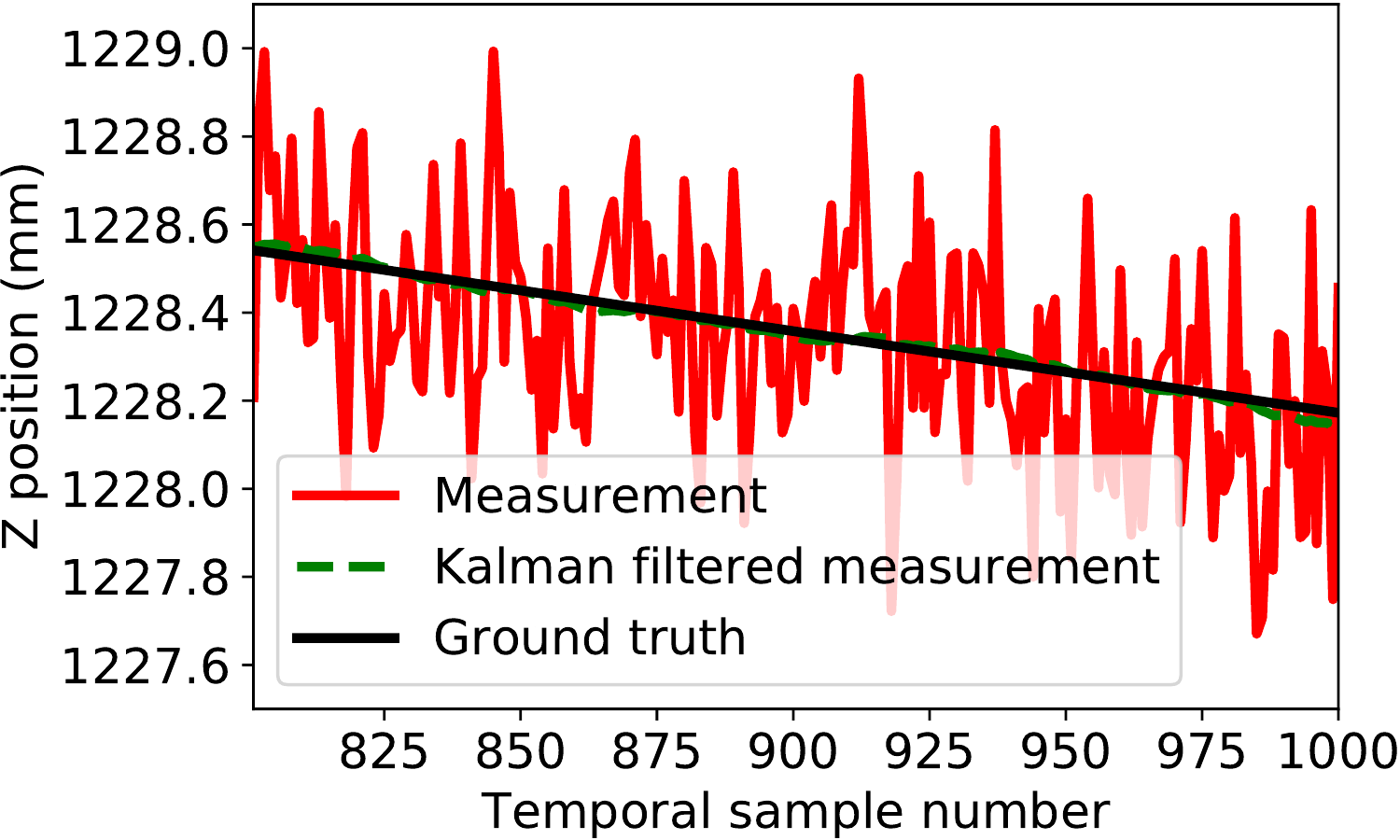} }}
	\caption{Temporal tracking plots for fiducial 1 of the simulated array. Columns 1-3 refer to $x$, $y$ and $z$ directions, respectively.}
	\label{plots_simulation}
\end{figure*}

\begin{figure*}
	%\DeclareGraphicsExtensions{.eps}
	\centering
	\subfigure[Fiducial 1, $x$ direction]{{\includegraphics[width=.33\textwidth]{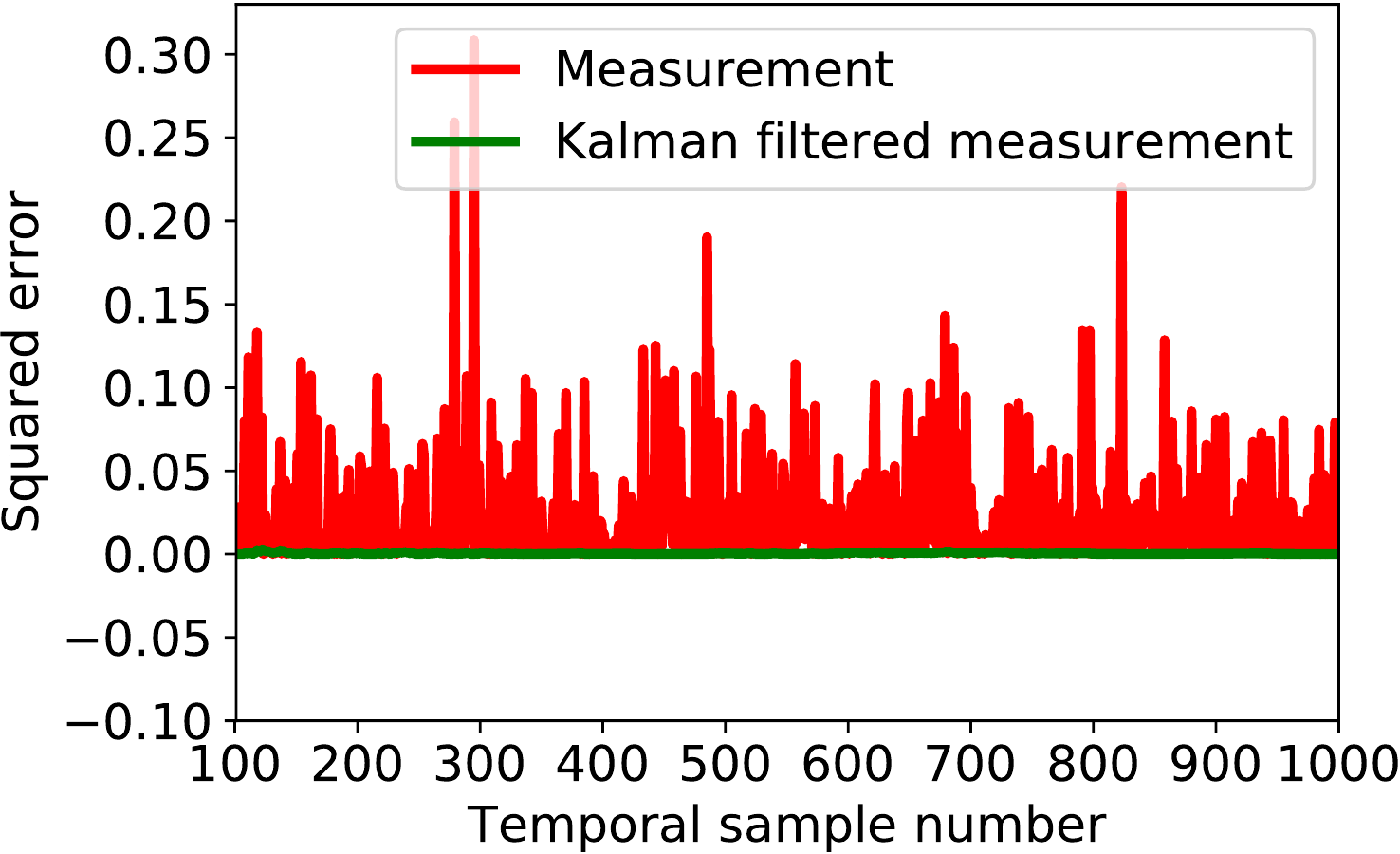}}}%
	\subfigure[Fiducial 1, $y$ direction]{{\includegraphics[width=.33\textwidth]{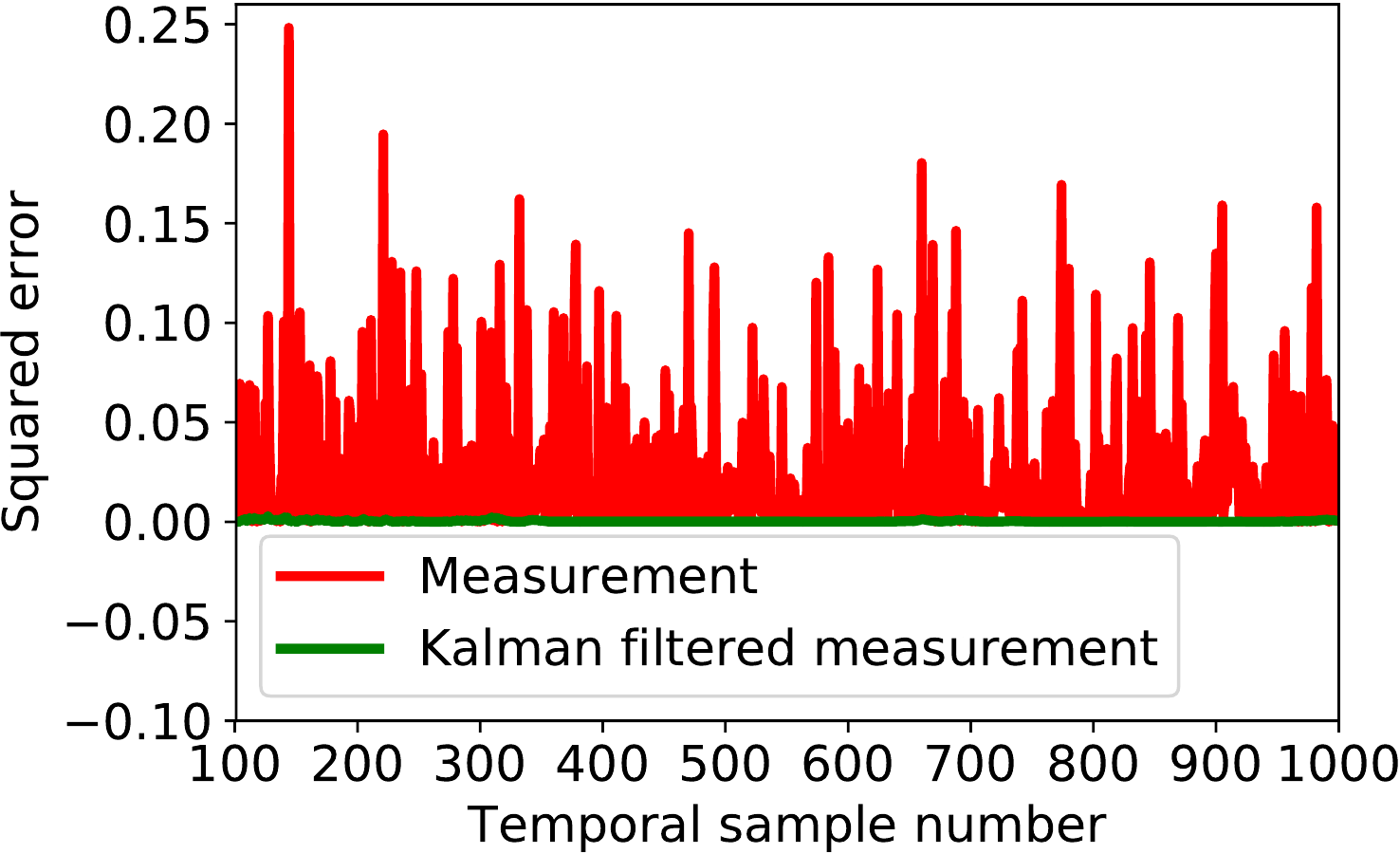}}}%
	\subfigure[Fiducial 1, $z$ direction]{{\includegraphics[width=.33\textwidth]{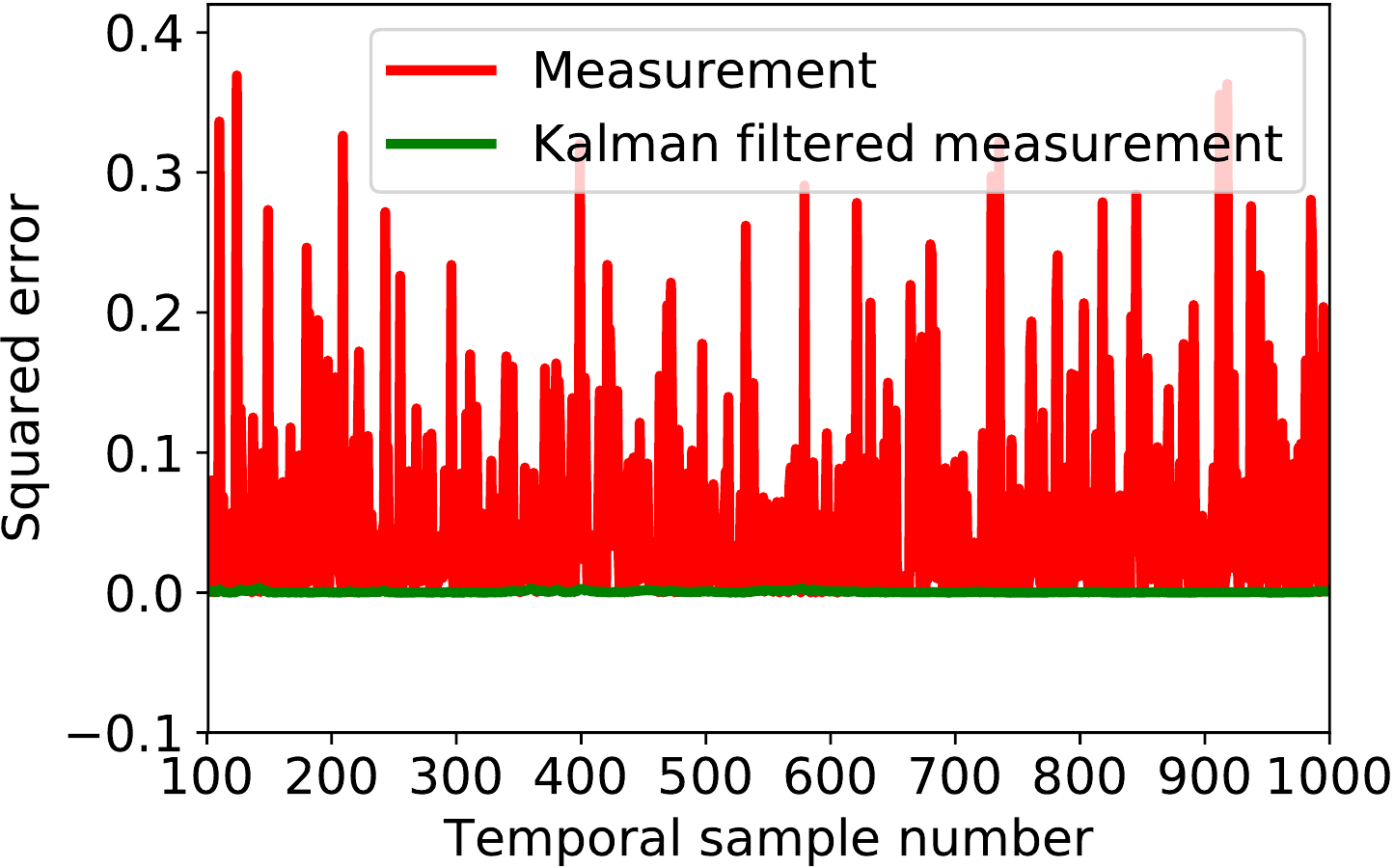} }}
	\caption{Squared error plots for fiducial 1 of the simulated array. Columns 1-3 refer to $x$, $y$ and $z$ directions, respectively.}
	\label{error_simulation}
\end{figure*}

The tracking results for one of the fiducials of the simulated array in all three directions are reported in Fig.~\ref{plots_simulation}. To maintain the conciseness, in the plots, we show the last 200 samples out of 1000 temporal samples obtained from 5 seconds of acquisition for only one fiducial. KF achieves a similar level of noise suppression in other fiducials. These results show that KF minimizes the measurement noise and substantially improves the tracking quality. In $x$ and $y$ directions, the filtered outputs exhibit almost no difference with the ground truth positions of the fiducials. Since the noise model is anisotropic and the variance is $40\%$ higher in the $z$ direction, the tracking error in this direction is slightly higher. Fig.~\ref{error_simulation} shows the square error plots for all fiducials in all three directions. Since KF requires some time in the beginning to adapt with the motion trajectory, we consider the first 0.5 second as the burnout period and therefore, disregard the first 100 temporal samples during error calculation. MSE and error variance values reported in Table~\ref{table_simulation} substantiate our visual assessment, showing a reduction in the tracking error from the order of $10^{-2}$~$mm^{2}$ to $10^{-4}$~$mm^{2}$.

\begin{table}[tb]  
	\centering
	\caption{Quantitative values of MSE and error varaince for the simulated dataset.}
	\label{table_simulation}
	\begin{tabular}{c c c c c c c c c c c c} 
		%\hline
		\cline{1-6} 
		\multicolumn{1}{c}{} &
		\multicolumn{2}{c}{MSE ($mm^{2}$)} &
		%\multicolumn{3}{c}{} &
		\multicolumn{1}{c}{} &
		\multicolumn{2}{c}{Error variance ($mm^{2}$)}\\
		%\multicolumn{3}{c}{}\\ 
		\cline{2-3} 
		\cline{5-6}
		$ $  $ $& $ $  $ $  Without KF  & With KF $ $  $ $&&$ $  $ $ $ $  $ $ Without KF & With KF\\
		\cline{1-6}
		Fiducial 1, X & $2.1 \times 10^{-2}$ & $3.33 \times 10^{-4}$ && $2.09 \times 10^{-2}$ & $2.47 \times 10^{-4}$ \\
		Fiducial 1, Y & $2.24 \times 10^{-2}$ & $2.51 \times 10^{-4}$ && $2.24 \times 10^{-2}$ & $2.41 \times 10^{-4}$ \\
		Fiducial 1, Z & $4.49 \times 10^{-2}$ & $4.87 \times 10^{-4}$ && $4.49 \times 10^{-2}$ & $4.39 \times 10^{-4}$\\
		Fiducial 2, X & $2.17 \times 10^{-2}$ & $7.51 \times 10^{-4}$ && $2.17 \times 10^{-2}$ & $6.49 \times 10^{-4}$ \\
		Fiducial 2, Y & $2.3 \times 10^{-2}$ & $2.35 \times 10^{-4}$ && $2.30 \times 10^{-2}$ & $2.35 \times 10^{-4}$ \\
		Fiducial 2, Z & $4.28 \times 10^{-2}$ & $9.46 \times 10^{-4}$ && $4.27 \times 10^{-2}$ & $5.67 \times 10^{-4}$\\
		Fiducial 3, X & $2.28 \times 10^{-2}$ & $8.07 \times 10^{-4}$ && $2.28 \times 10^{-2}$ & $7.22 \times 10^{-4}$ \\
		Fiducial 3, Y & $2.25 \times 10^{-2}$ & $4.23 \times 10^{-4}$ && $2.25 \times 10^{-2}$ & $4.2 \times 10^{-4}$ \\
		Fiducial 3, Z & $4.32 \times 10^{-2}$ & $8.44 \times 10^{-4}$ && $4.32 \times 10^{-2}$ & $7.62 \times 10^{-4}$\\
		Fiducial 4, X & $2.36 \times 10^{-2}$ & $5.05 \times 10^{-4}$ && $2.35 \times 10^{-2}$ & $3.69 \times 10^{-4}$ \\
		Fiducial 4, Y & $2.02 \times 10^{-2}$ & $3.87 \times 10^{-4}$ && $2.02 \times 10^{-2}$ & $3.66 \times 10^{-4}$ \\
		Fiducial 4, Z & $4.14 \times 10^{-2}$ & $4.5 \times 10^{-4}$ && $4.13 \times 10^{-2}$ & $4.43 \times 10^{-4}$\\
		\cline{1-6}
	\end{tabular}
\end{table}

\subsection{Real dataset}
\begin{figure*}
	%\DeclareGraphicsExtensions{.eps}
	\centering
	\subfigure[Fiducial 1, $x$ direction]{{\includegraphics[width=.33\textwidth]{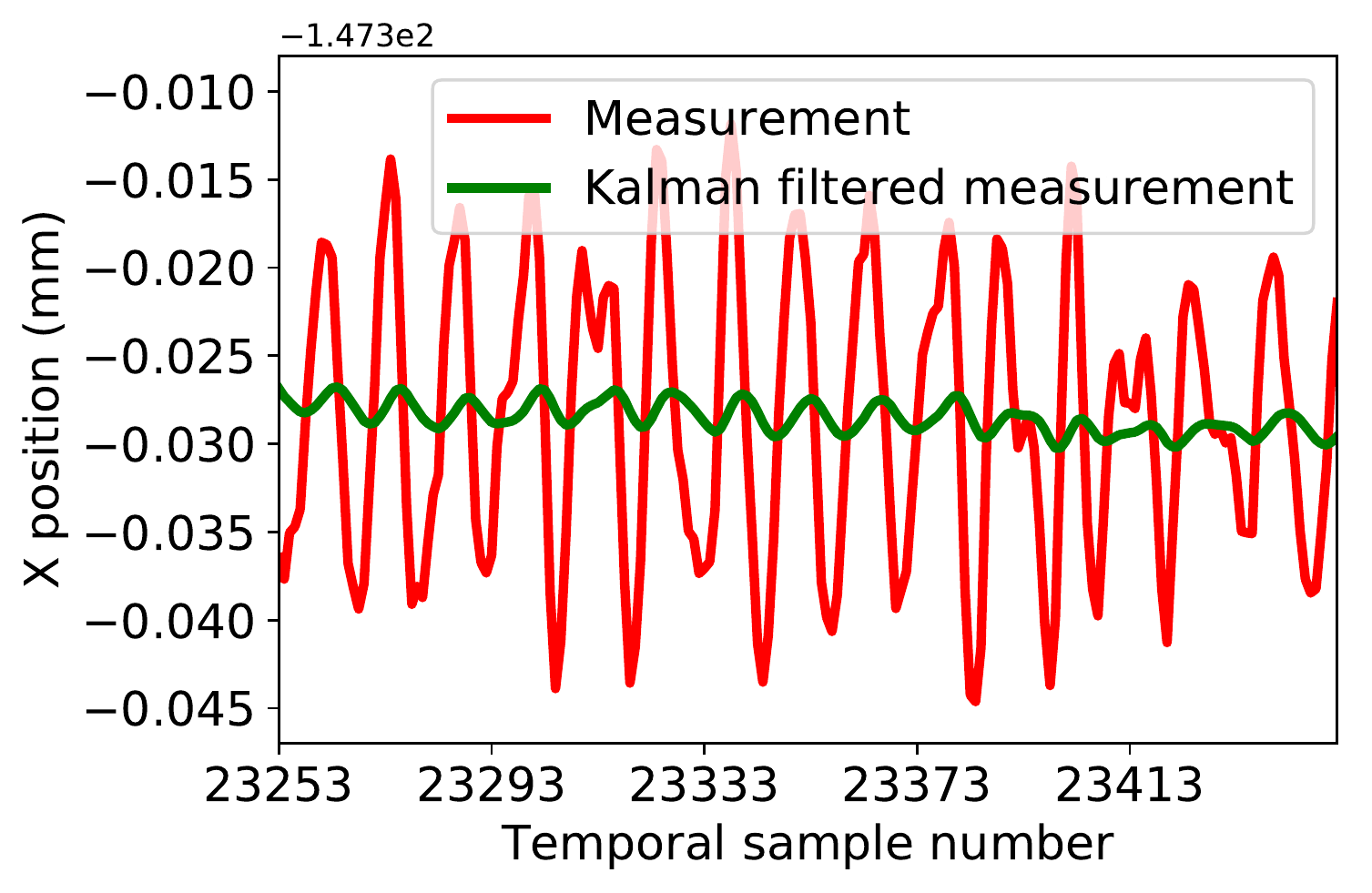}}}%
	\subfigure[Fiducial 1, $y$ direction]{{\includegraphics[width=.33\textwidth]{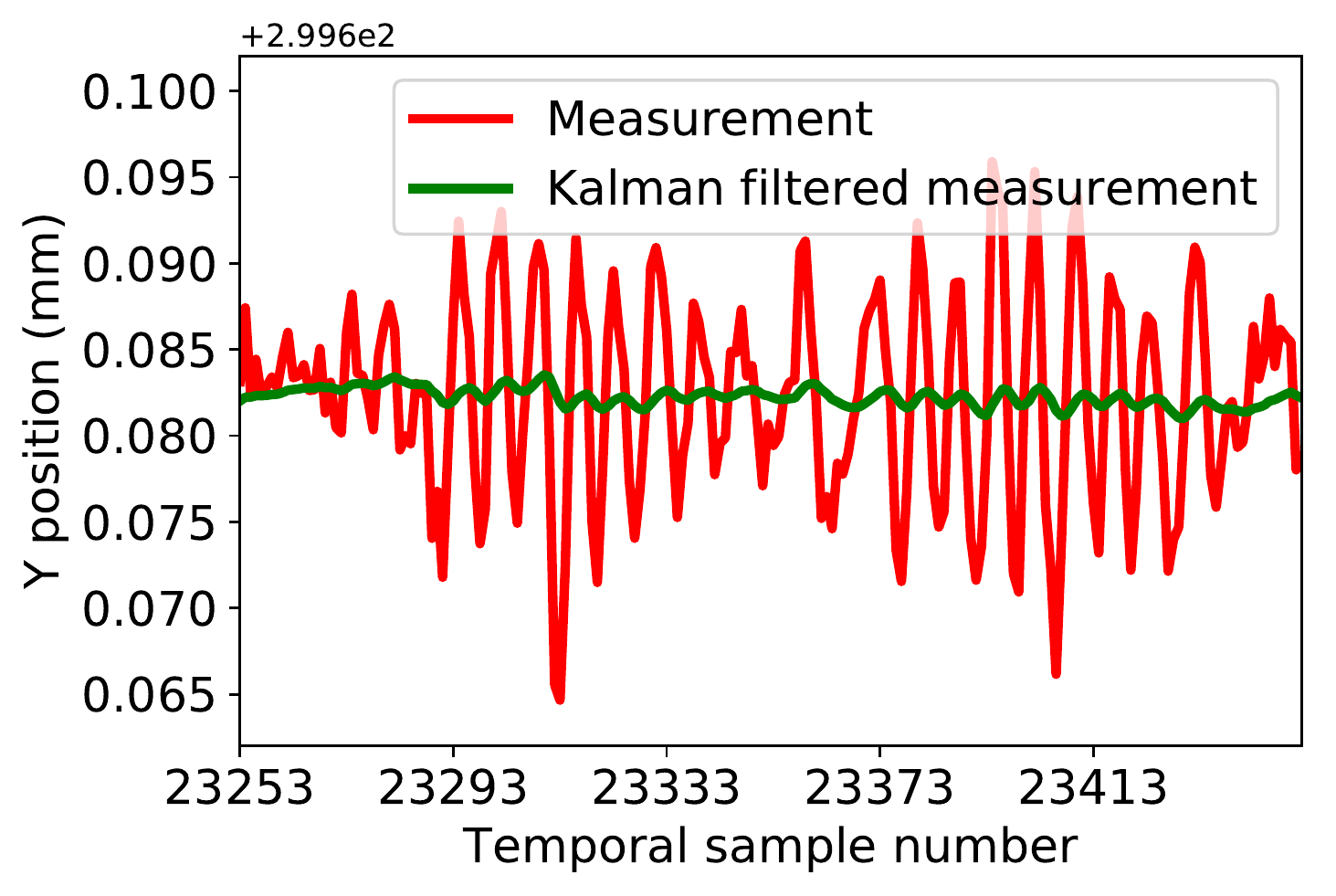}}}%
	\subfigure[Fiducial 1, $z$ direction]{{\includegraphics[width=.33\textwidth]{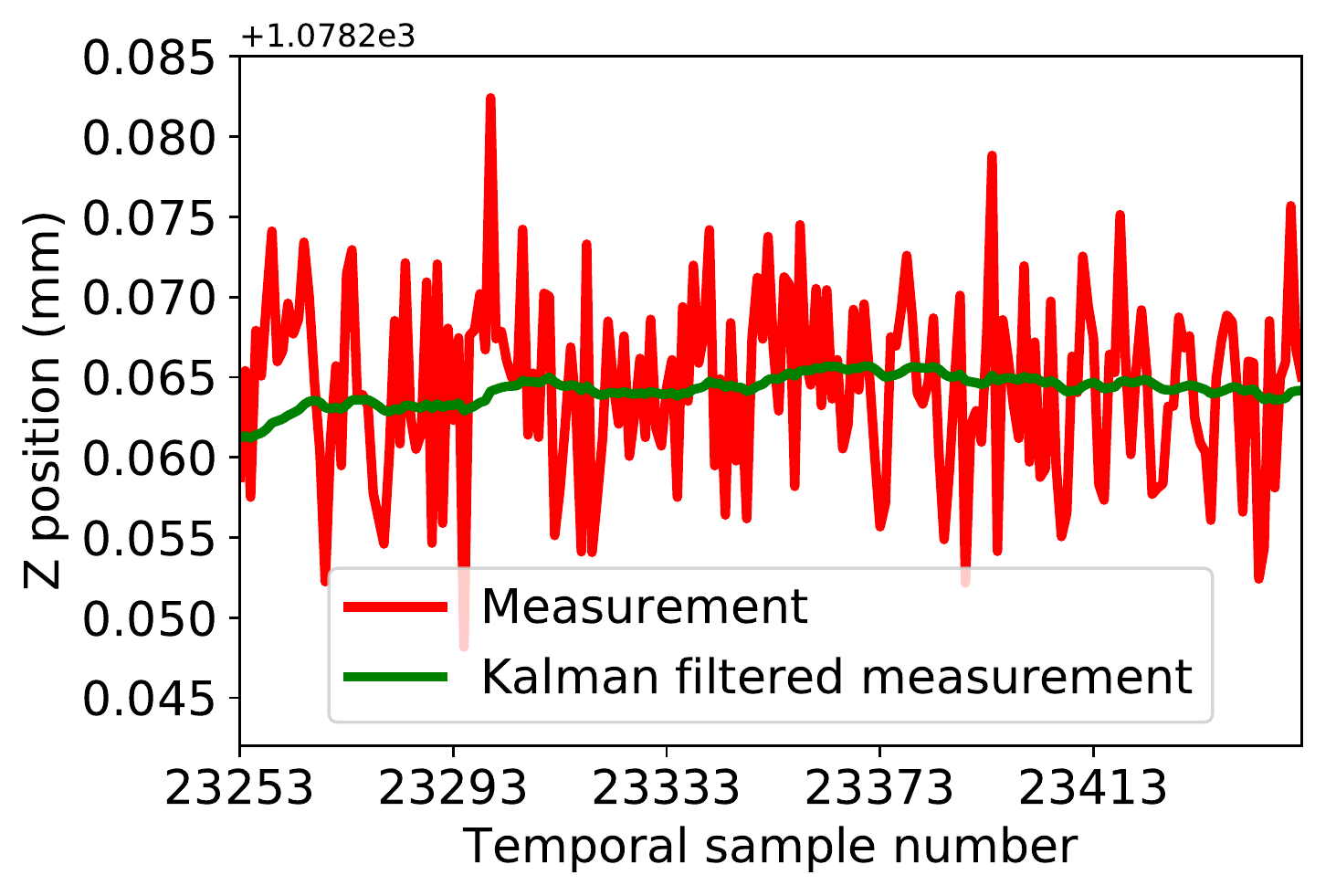} }}
	\caption{Temporal tracking plots for the real dataset collected from static marker. Columns 1-3 refer to $x$, $y$ and $z$ directions, respectively.}
	\label{plots_real_static}
\end{figure*}

Our first experiment analyzes the performance with the dataset acquired from the static digitizer, where the ground truth velocity is zero. Since this dataset is collected from a steady marker, position constancy of all fiducials in all three directions is expected. Fig.~\ref{plots_real_static} shows that  current tracking system exhibits extensive variation around the expected positions. It is evident from this figure that combining KF with the existing system resolves the issue of measurement variation by stabilizing the position tracking in all three directions.    

\begin{figure*}
	%\DeclareGraphicsExtensions{.eps}
	\centering
	\subfigure[Fiducial 1, $x$ direction]{{\includegraphics[width=.33\textwidth]{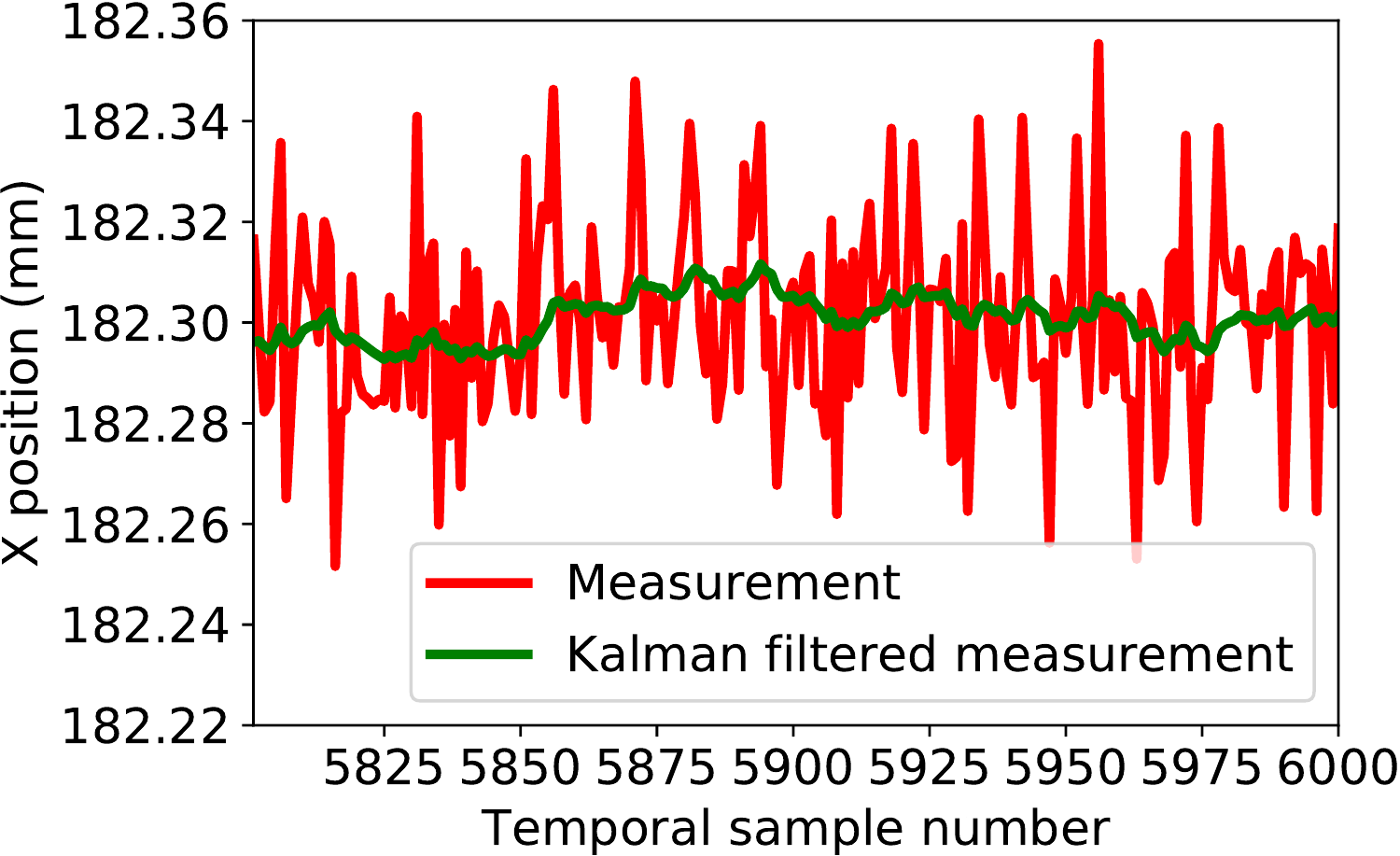}}}%
	\subfigure[Fiducial 1, $y$ direction]{{\includegraphics[width=.33\textwidth]{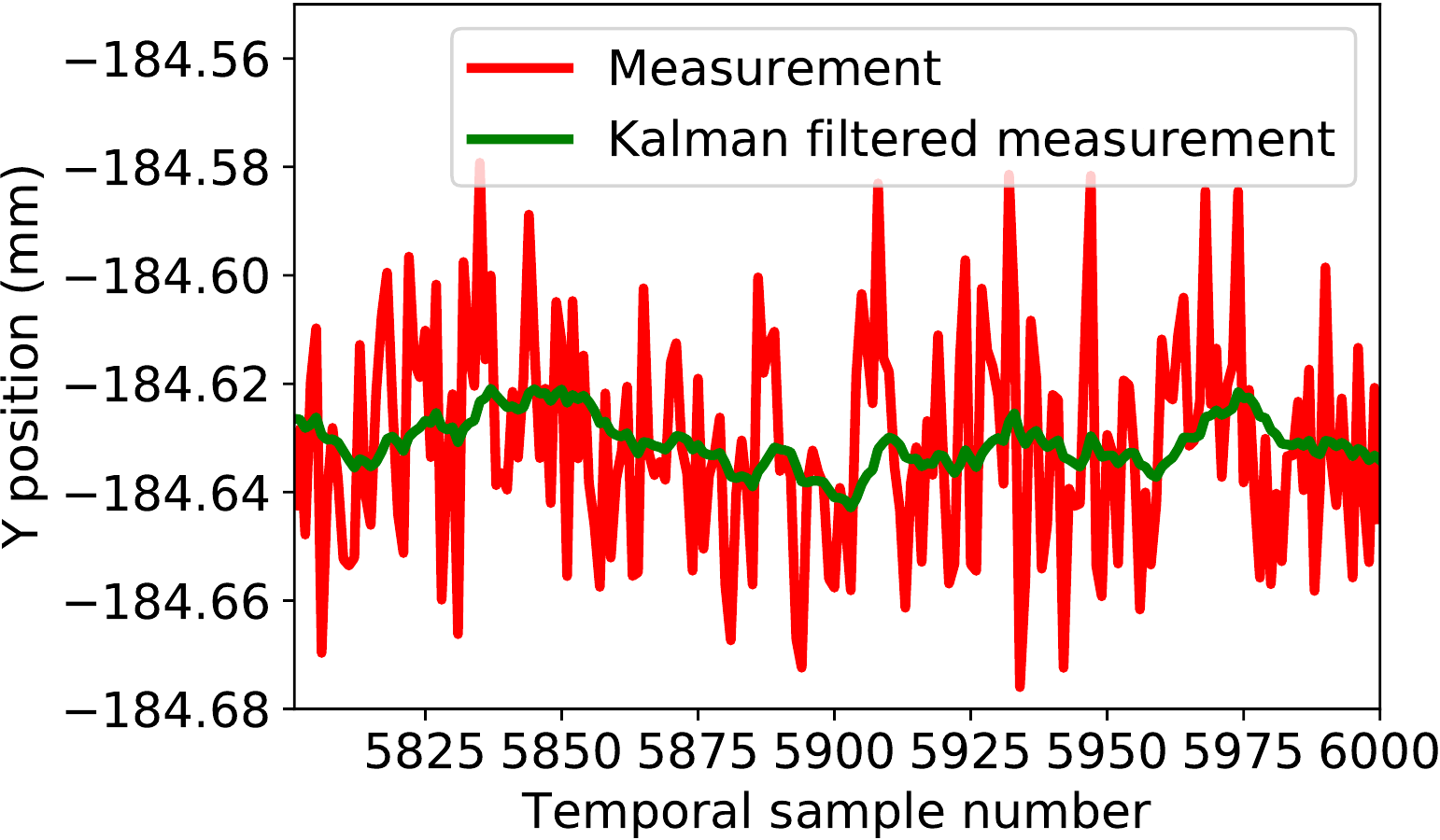}}}%
	\subfigure[Fiducial 1, $z$ direction]{{\includegraphics[width=.33\textwidth]{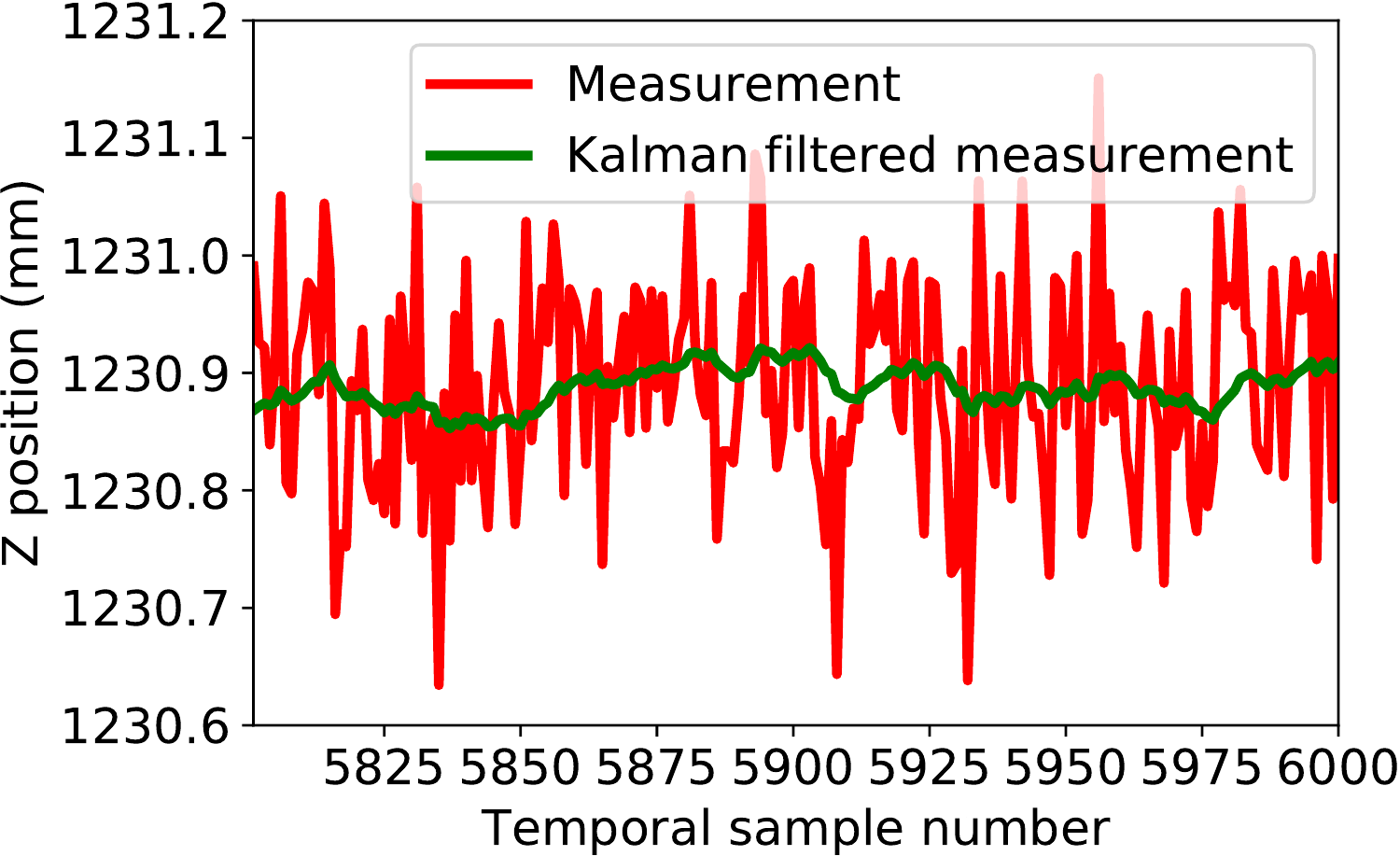} }}
	\caption{Temporal tracking plots for array undergoing rotational and translational motion. Columns 1-3 correspond to $x$, $y$ and $z$ directions, respectively.}
	\label{plots_real}
\end{figure*}

In the second experiment, we investigate the tracking performance of the marker that undergoes rotational and translational motion. The results of noisy measurement of the positions of one of the fiducials along with the KF measurements are presented in Fig.~\ref{plots_real}. Like the previous experiment, we show the last 200 temporal samples out of a total of 6000 samples to keep the figures comprehensible. The position plots show that the Kalman filtered outputs manifest substantially lower fluctuation compared to the raw measurement. It is observed that KF noticeably stabilizes the measurements of the first fiducial and slight variations around 182.3 mm, -184.63 mm, and 1230.88 mm are observed in $x$, $y$ and $z$ directions, respectively. Similar tracking performance is achieved for the other three fiducials as well.

\begin{figure*}
	%\DeclareGraphicsExtensions{.eps}
	\centering
	\subfigure[Fiducial 1, $x$ direction]{{\includegraphics[width=.33\textwidth]{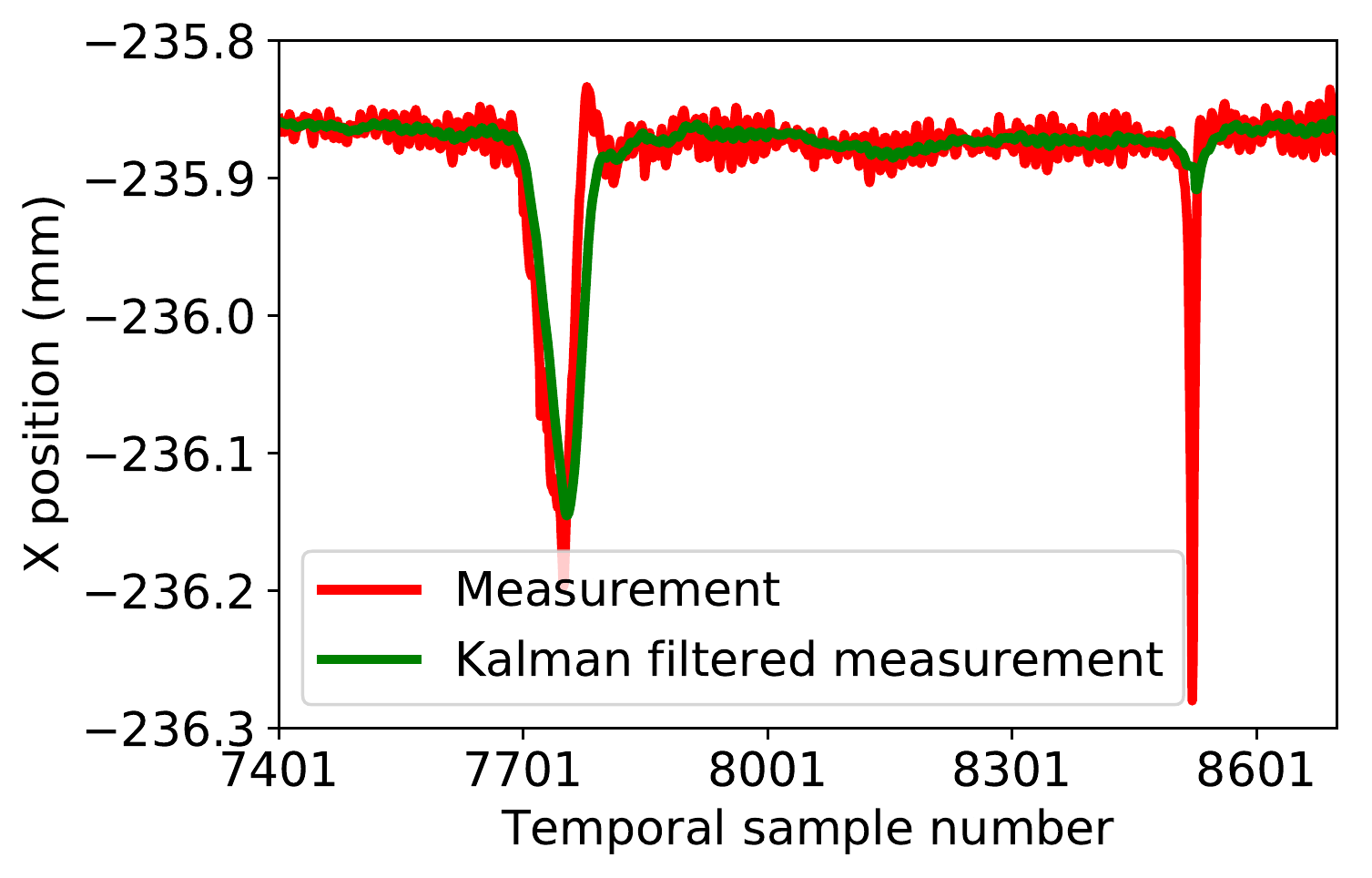}}}%
	\subfigure[Fiducial 1, $y$ direction]{{\includegraphics[width=.33\textwidth]{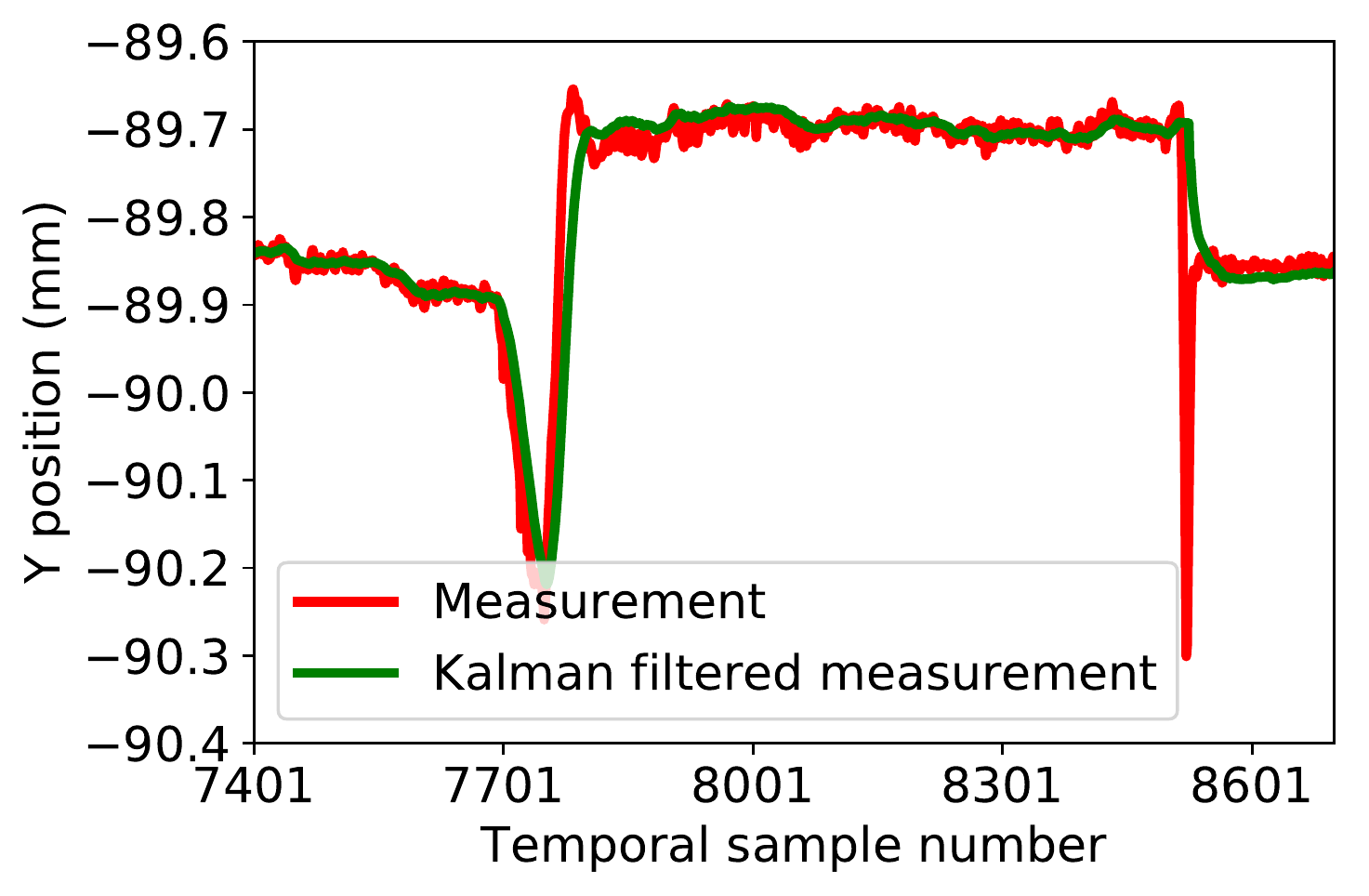}}}%
	\subfigure[Fiducial 1, $z$ direction]{{\includegraphics[width=.33\textwidth]{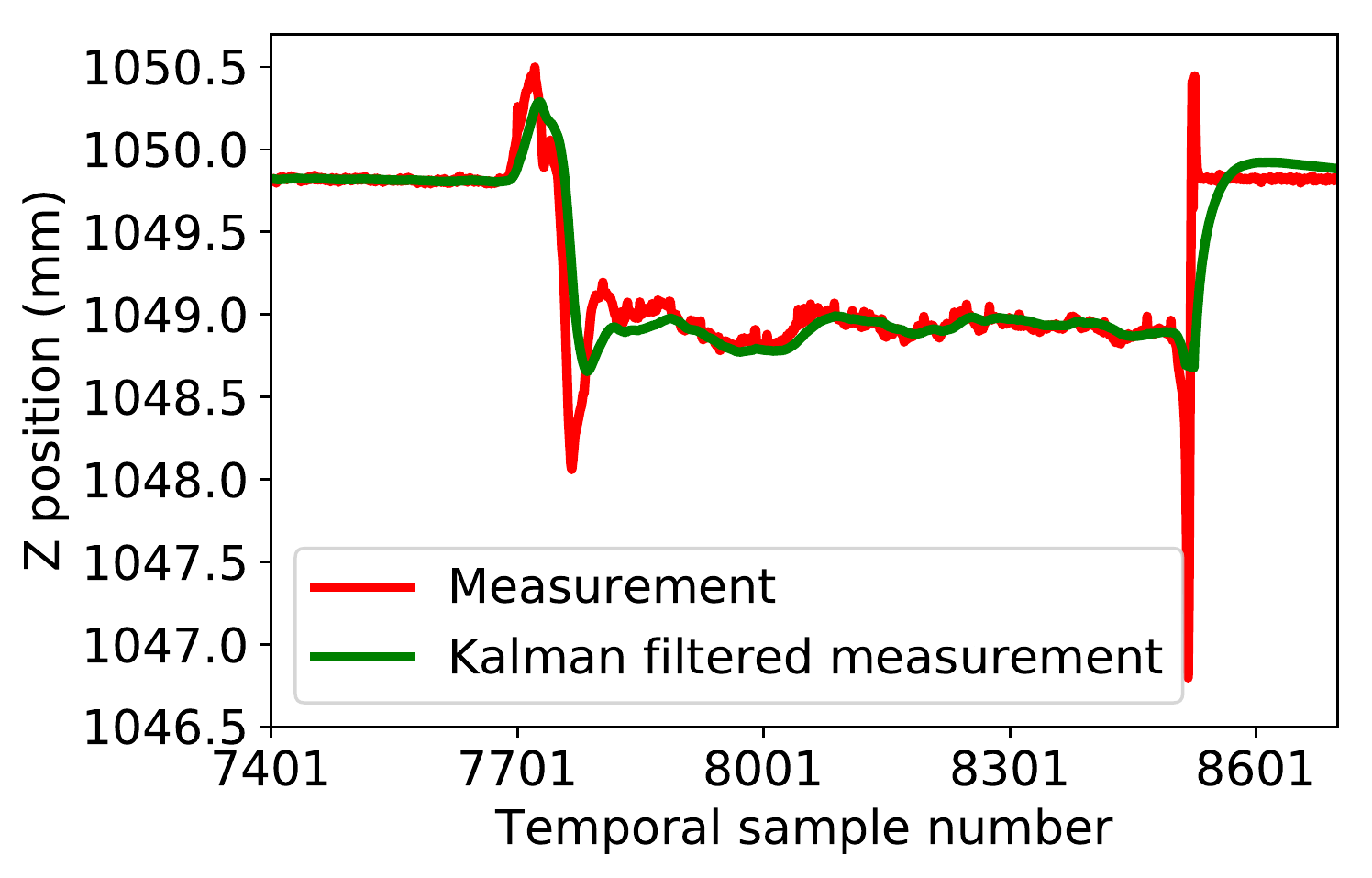} }}
	\subfigure[Fiducial 3, $x$ direction]{{\includegraphics[width=.33\textwidth]{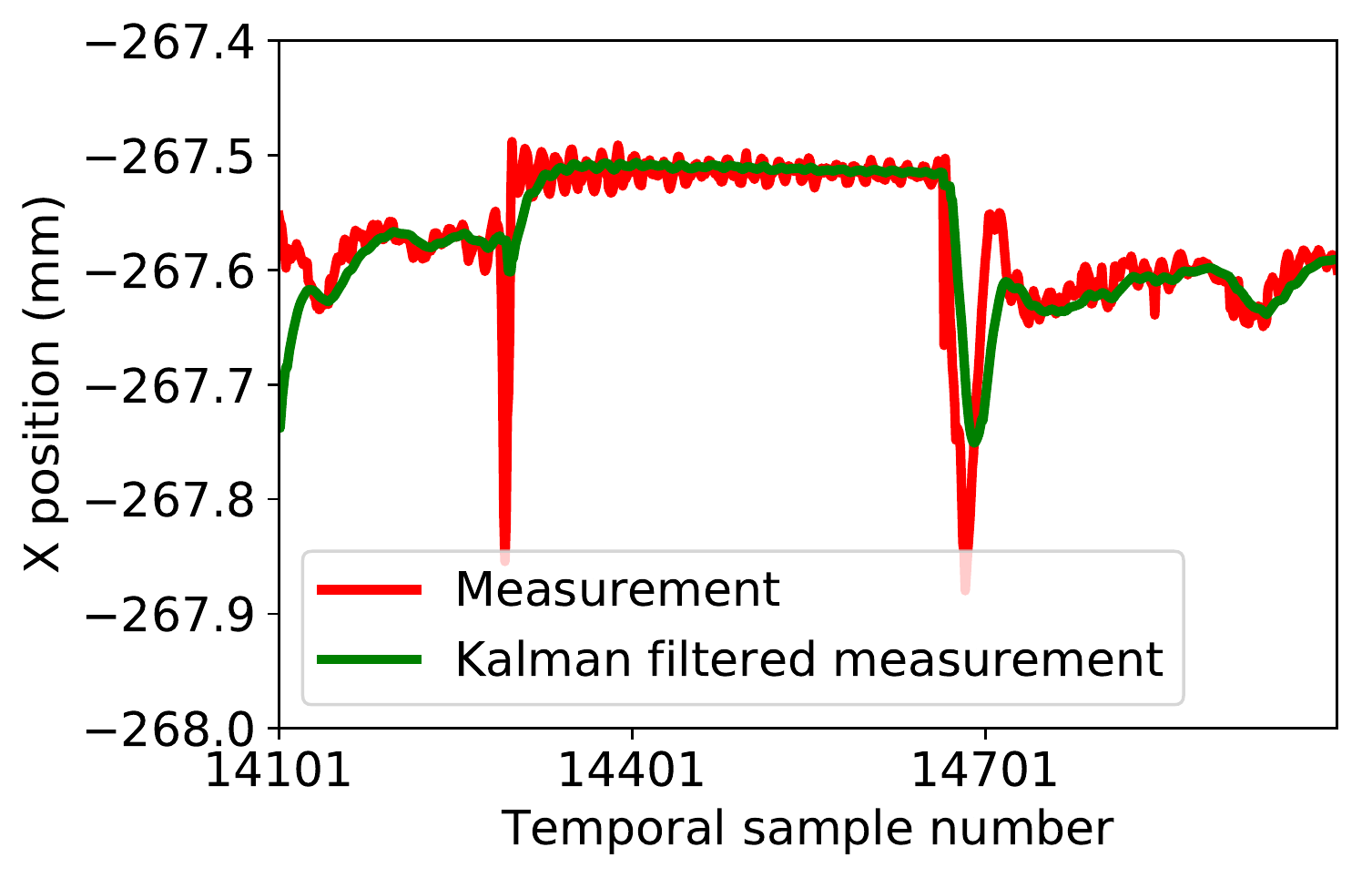}}}%
	\subfigure[Fiducial 3, $y$ direction]{{\includegraphics[width=.33\textwidth]{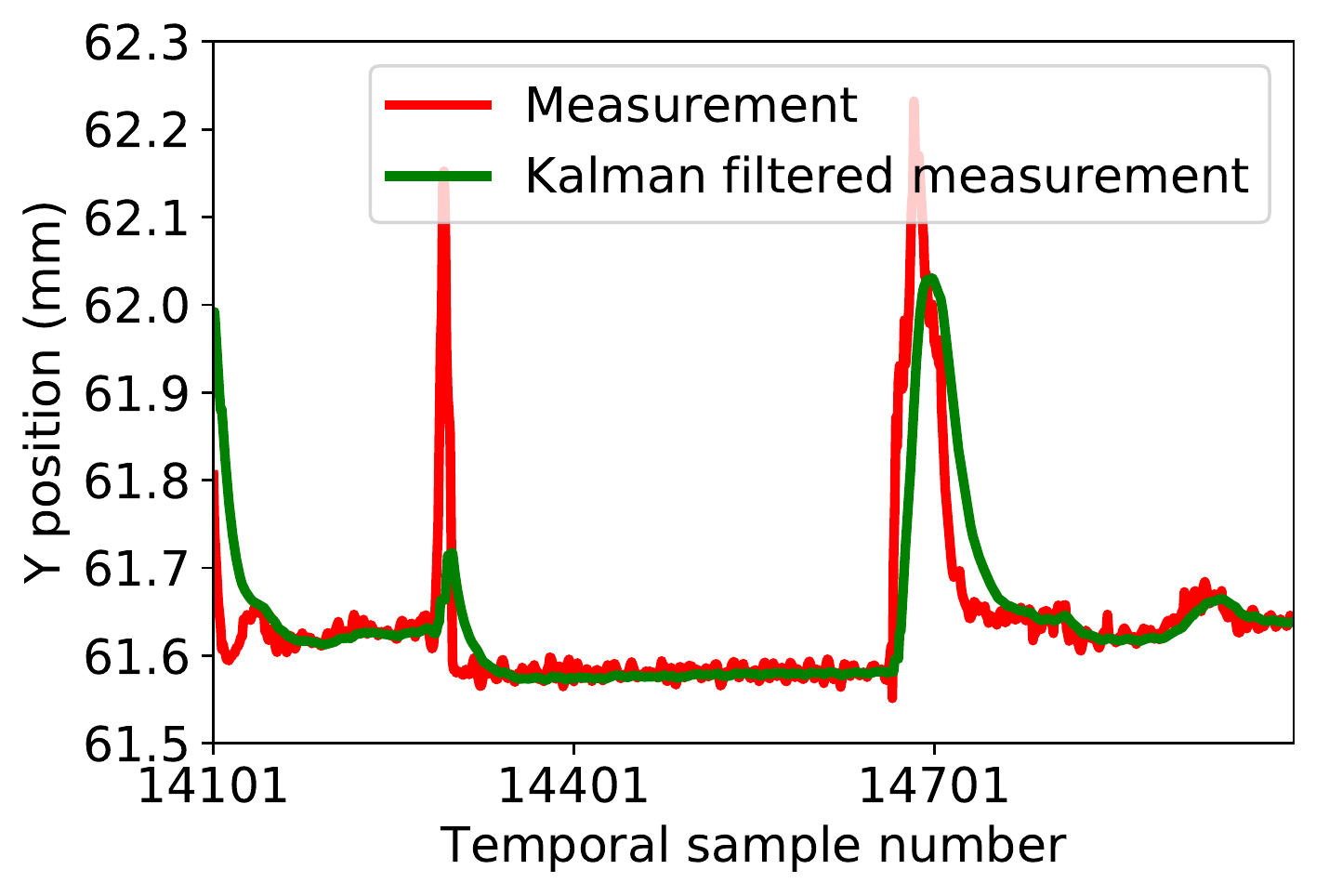} }}%
	\subfigure[Fiducial 3, $z$ direction]{{\includegraphics[width=.33\textwidth]{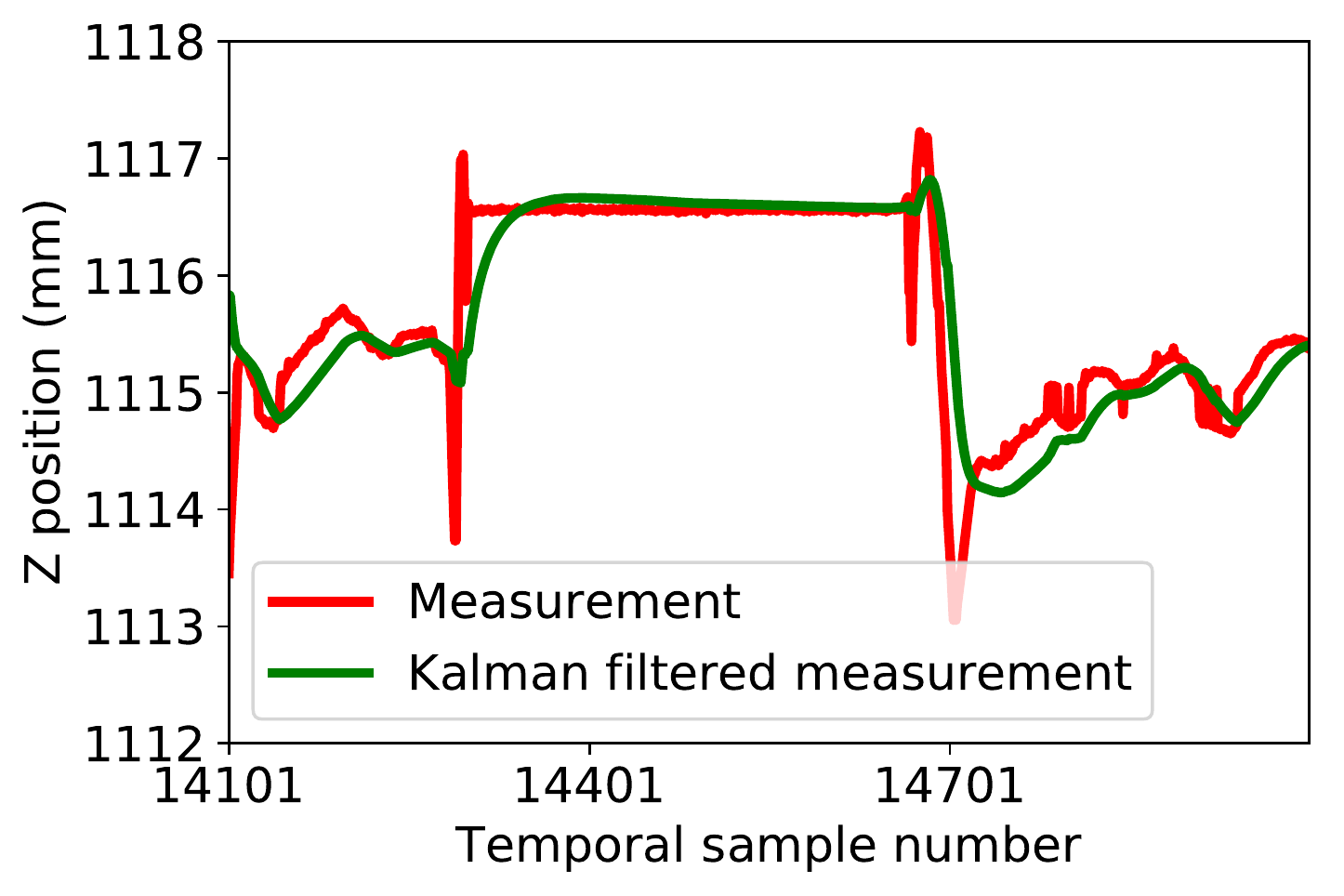} }}
	\caption{Tracking results for blocked fiducials. Rows 1 and 2 correspond to fiducials 1 and 3, respectively whereas columns 1-3 refer to $x$, $y$ and $z$ directions, respectively.}
	\label{plots_real_occu}
\end{figure*}

The third experiment examines the performance of the proposed KF when the fiducial of interest is blocked by a translucent material. Fig.~\ref{plots_real_occu} reports the temporal regions where fiducials 1 and 3 are blocked. In all three directions of both fiducials, extensive discontinuities are observed at the instants of disposal and removal of the glass diffuser. However, during the stable placement of diffuser, the first fiducial's $y$ and $z$ positions are overestimated and underestimated, respectively whereas the $x$ position remains unaffected. In case of third fiducial, the $z$ measurement exhibits large upward shift whereas slight overestimation and underestimation are noticed in $x$ and $y$ directions, respectively. In all cases, the output of KF follows the trend of the actual measurement, but with substantially lower variance. Besides, the proposed scheme successfully suppresses the spurious spikes introduced by diffuser placement and withdrawal.        

\section{Discussion and conclusion}
As reported in Fig.~\ref{plots_real_occu}, the proposed implementation of KF follows the trend of the incorrect position measurement when any of the fiducials moves out of the field-of-view or gets blocked by a translucent material such as a drop of blood. Although it reduces the estimation variance, it cannot amend the step-like over or underestimation of fiducial position, likely caused by light diffraction. The rigid-body model incorporated in EKF and UKF implementations empowers the system to adapt with the situation of fiducial occlusion and reconstruct the surgical tool with modest tracking error. However, since this scheme tracks the instrument as a whole, it cannot notify the surgeon which of the fiducials is blocked or out of field-of-view (FOV). This drawback can potentially be resolved with the proposed KF framework.        

Herein, we proposed a fast implementation of linear KF on a high frame-rate tracking system where a Newtonian model was taken into account to track each fiducial of a surgical tool individually. Besides facilitating real-time surgical tracking, this technique efficiently suppresses acquisition and estimation noise experienced by an optical tracking system. In addition, high performance in dynamic localization of intraoperative instruments proves that the proposed framework eliminates the requirement of rigid-body constraint while tracking a surgical tool at high temporal resolution.

\begin{acknowledgements}
The authors acknowledge funding from Natural Science and Engineering Research Council of Canada (NSERC). 
\end{acknowledgements}

\section*{Conflict of interest}
The authors declare that they have no conflict of interest.

%\bibliographystyle{spmpsci}
%\bibliographystyle{spphys}
\bibliographystyle{spmpsci}
\bibliography{IPCAI_THINK}

\end{document}